\documentclass{article}

\PassOptionsToPackage{numbers, compress}{natbib}
\usepackage[main,final]{neurips_2026}

\makeatletter
\renewcommand{\@noticestring}{}
\makeatother

\usepackage[utf8]{inputenc}
\usepackage[T1]{fontenc}
\usepackage{hyperref}
\usepackage{url}
\usepackage{booktabs}
\usepackage{amsfonts}
\usepackage{nicefrac}
\usepackage{microtype}
\usepackage{xcolor}
\usepackage{amsmath}
\usepackage{amssymb}
\usepackage{graphicx}
\usepackage{multirow}
\usepackage{makecell}
\usepackage{array}
\usepackage{colortbl}

\usepackage{tikz}
\usepackage{pgfplots}
\pgfplotsset{compat=1.18}
\usetikzlibrary{
  arrows.meta,
  positioning,
  shapes.geometric,
  shapes.misc,
  calc,
  fit,
  decorations.pathreplacing,
  backgrounds,
  patterns
}

\definecolor{myblue}{RGB}{31,119,180}
\definecolor{myorange}{RGB}{255,127,14}
\definecolor{mygreen}{RGB}{44,160,44}
\definecolor{myred}{RGB}{214,39,40}
\definecolor{mygray}{RGB}{127,127,127}
\definecolor{mypurple}{RGB}{148,103,189}
\definecolor{lightgray}{RGB}{240,240,240}
\definecolor{darkblue}{RGB}{8,48,107}
\definecolor{goodgreen}{RGB}{0,109,44}
\definecolor{badred}{RGB}{165,15,21}

\title{Two Calls Beat Five Agents: Evaluating Multi-Agent Pipelines\\Against Self-Refinement for Local Language Models}

\author{%
  Ashish Prajapati \\
  Independent Researcher \\
  Mumbai, India \\
  \texttt{ashish.prajapati.research@gmail.com} \\
  \And
  Om Mohite \\
  Independent Researcher \\
  Mumbai, India \\
  \texttt{om.mohite.research@gmail.com}
}

\begin{document}

\maketitle

\begin{abstract}
Multi-agent LLM pipeline systems break down the task among multiple roles for better reasoning, but are benchmarked mainly with large-scale commercial models. In this study, we investigate Parishad, a structured multi-agent system involving five roles, by deploying it on Qwen2.5-7B-Instruct, a local model, on two datasets: GSM8K (500 questions) and HumanEval (164 questions), compared with prompting directly and two-call self-refinement. The multi-agent system drops GSM8K accuracy from 75.0\% to 45.0\% with JSON data format due to the error accumulation problem. With plaintext format, the accuracy is restored to 82.0\%. A two-call self-refinement strategy (V1) can achieve 86.2\% accuracy on GSM8K, with 7.4$\times$ lower token usage. However, the same V1 implementation on HumanEval---where direct accuracy is already 96.3\%---actively destroys performance (66.5\%). A task-aware gated redesign (V2) applied to HumanEval preserves accuracy at 95.1\%. Our results demonstrate that communication format and implementation details determine outcomes more than architectural complexity, and that simpler approaches match or outperform multi-agent pipelines for local 7B model deployment. All code and data are released.
\end{abstract}

%==========================================================================
\section{Introduction}
%==========================================================================

Multi-agent LLM systems have emerged as a dominant paradigm for complex reasoning tasks. Systems such as ChatDev~\cite{qian2024chatdev}, MetaGPT~\cite{hong2024metagpt}, AutoGen~\cite{wu2023autogen}, and CrewAI~\cite{crewai2024} decompose problems across multiple specialized agents, each contributing a distinct capability---planning, execution, verification, synthesis. These systems consistently show improvements over single-model baselines, often achieving state-of-the-art results on coding, reasoning, and software engineering benchmarks.

However, nearly all published evaluations of multi-agent systems use large commercial models: GPT-4, Claude, or Gemini. There is growing practical interest in running LLM systems locally on consumer hardware, for reasons of cost, privacy, and offline deployment. It is not established whether multi-agent architectures that help GPT-4 also help smaller local models. The architectural assumptions that hold for 100B+ parameter models---reliable instruction following, consistent structured output generation, stable context maintenance across long prompts---may not hold for 7B parameter models running on consumer hardware.

We evaluate this question directly. We take Parishad, a structured five-role pipeline (Refiner $\to$ Planner $\to$ Worker $\to$ Checker $\to$ Judge), and run it against direct prompting and two-call self-refinement on a local Qwen2.5-7B-Instruct model. We test on GSM8K~\cite{cobbe2021gsm8k} for mathematical reasoning and HumanEval~\cite{chen2021humaneval} for code generation. Our findings are surprising in their clarity.

Our contributions are:
\begin{enumerate}
  \item The first empirical evaluation of a structured multi-role pipeline against self-refinement on a local 7B model across both math and code benchmarks.
  \item Identification of JSON inter-role communication as the primary failure mechanism in local model pipelines, with a 37 percentage-point recovery from switching to plain text.
  \item Evidence that two-call self-refinement outperforms the fixed pipeline on math ($+$4.2\%) using 7.4$\times$ fewer tokens.
  \item A systematic ablation showing that self-refinement implementation details---confidence gating and prompt design---determine whether refinement helps ($+$11.2\%) or actively destroys performance ($-$29.8\%) on code tasks.
  \item A practical design recommendation: for local 7B model deployment, communication format and task-aware implementation matter more than architectural complexity.
\end{enumerate}

%==========================================================================
\section{Related Work}
%==========================================================================

\subsection{Multi-Agent LLM Systems}

ChatDev~\cite{qian2024chatdev} and MetaGPT~\cite{hong2024metagpt} pioneered the use of role-based multi-agent systems for software development, assigning agents to roles such as programmer, reviewer, and tester. The AutoGen system~\cite{wu2023autogen} proposes a generic architecture for multi-agent communication. The AgentCoder system~\cite{huang2023agentcoder} utilizes multi-agent coordination for code synthesis. Both systems show consistent improvement over single-agent systems. In particular, MetaGPT demonstrates considerable performance improvement on HumanEval benchmarks. Nevertheless, all these systems are benchmarked using GPT-3.5 or GPT-4 class models.

\subsection{Self-Refinement}

Self-Refine is a system proposed by Madaan et al.~\cite{madaan2023selfrefine}, wherein a model repeatedly improves its generated outputs based on feedback from a critic. It achieves stable gains on various tasks such as mathematics, programming, and dialogue. Most importantly, it was extensively tested using GPT-3.5 and GPT-4. On the other hand, Shinn et al.~\cite{shinn2023reflexion} introduce Reflexion, which uses verbal reinforcement to enhance the decisions made by an agent over several trials. Both methods depend on the capability of the model to evaluate itself---which we validate in our experiment.

\subsection{Small Model Reasoning}

Recent work wanted to see what smaller models are able to do. They explored the limits of reasoning in these models. Wei et al.~\cite{wei2022chain} showed that chain-of-thought prompting only reliably emerges in models above approximately 100B parameters, though subsequent work found earlier emergence with instruction tuning. In addition to Kojima et al.~\cite{kojima2022large}, who demonstrated that ``Let's think step by step'' can be transferred to small models, our paper contributes to this body of work by studying whether multi-step architectural patterns, but not prompting patterns, transfer to 7B models.

\subsection{Gap Addressed}

None of the existing literature has made a direct comparison between multi-role pipelines and self-refinement on sub-10B models. Our evaluation closes this gap through rigorous experimental comparison using the same model, hardware, and benchmark tasks.

%==========================================================================
\section{System and Methods}
%==========================================================================

\subsection{The Parishad Pipeline}

Parishad is an organized five-role architecture for orchestrating LLMs. It follows a pipeline where the output of one task serves as context for the subsequent task within the pipeline. The five roles are:

\textbf{Darbari (Refiner):} Receives the raw user query and produces a structured task specification identifying the problem type, constraints, difficulty, and output format.

\textbf{Majumdar (Planner):} Receives the task specification and produces a step-by-step execution plan.

\textbf{Sainik (Worker):} Receives the query, task specification, and plan, and generates the primary solution---code, computation, or text.

\textbf{Prerak (Checker):} Receives all previous outputs and validates the Worker's solution for correctness, completeness, and consistency.

\textbf{Raja (Judge):} Receives all outputs and synthesizes a final authoritative answer, resolving any conflicts between the Worker and Checker.

\subsection{Communication Variants}

\textbf{Pipeline V1 (JSON):} Each role is instructed to output a structured JSON object with specific fields. The engine parses JSON from each role's output and passes the parsed structure to the next role. When JSON parsing fails, the raw output text is passed downstream.

\textbf{Pipeline V2 (Plain Text):} Role system prompts are rewritten to elicit plain text output instead of JSON. Each role describes its analysis, plan, or solution in natural language. The engine passes raw text strings between roles with no parsing step. Role prompts include explicit formatting cues (numbered steps, labeled sections) but require no JSON structure.

\subsection{Self-Refinement}
\label{sec:sr_methods}

We implement two variants of two-call self-refinement. Both variants use the same Call~1 prompt. They differ only in how Call~2 is prompted and whether it is gated.

\textbf{Self-Refine V1 (Ungated, Generic):} Call~1 generates an initial solution using a task-appropriate system prompt. Call~2 runs unconditionally on all problems using a generic verification prompt that asks the model to check for errors and always provide the final answer in the appropriate format.V1 is tested on both GSM8K (500 questions) and HumanEval (164 questions). V1 succeeds on GSM8K but not HumanEval, which leads to the following redesign.

\textbf{Self-Refine V2 (Gated, Task-Aware):} Developed precisely to resolve V1's shortcomings with HumanEval. Call~1 is identical to V1. Call~2 is gated by a task-specific confidence check. For code tasks, the gate checks whether the response contains a complete closed code block; if so, Call~2 is skipped. For math tasks, the gate checks for uncertainty markers. The Call~2 prompt for code instructs the model to respond \texttt{CORRECT} if the solution is correct, and only rewrite if a specific concrete bug is identified. We evaluate V2 on HumanEval only (164 problems), since V1 already produced strong results on GSM8K.

\subsection{Direct Baseline}

A single model call with a task-appropriate system prompt. For math: ``Solve step by step. End with: The answer is [number].'' For code: ``Write a complete correct Python function. Wrap in \texttt{```python```} blocks.''

\subsection{Experimental Setup}

All experiments use Qwen2.5-7B-Instruct~\cite{bai2025qwen} in GGUF Q4\_K\_M quantization via \texttt{llama.cpp}, with a 4096-token context window and all GPU layers offloaded. Hardware: Windows, Intel Core i7 (Gen 12), 31GB RAM. Random seed: 42. GSM8K is evaluated on 500 test problems (Direct, Pipeline V2, Self-Refine V1); 200 problems for Pipeline V1. HumanEval uses all 164 problems for all methods. GSM8K accuracy is measured via regex-based numeric extraction; HumanEval uses pass@1 against provided unit tests.

%==========================================================================
\section{Results}
%==========================================================================

\subsection{Main Results}

Table~\ref{tab:main} shows the main comparison across methods and benchmarks. There are three clear trends.

\textbf{Pipeline V1 fails catastrophically compared to prompt-based reasoning.} Pipeline V1 underperforms by 30 percentage points on GSM8K and 40 percentage points on HumanEval relative to prompting directly on the problem. It appears that the entire pipeline architecture actually hinders reasoning abilities of the model.

\textbf{Plaintext communication almost fully compensates for the loss.} Changing JSON to plaintext (Pipeline V1 $\to$ V2) results in 37-point accuracy increase on GSM8K (45.0\% $\to$ 82.0\%) and large accuracy boost on HumanEval (56.0\% $\to$ 81.7\%). No changes have been made to the architecture, roles, model. Only the communication format was modified. This is the key insight of our paper: \emph{communication format, rather than architecture, is the key factor in local model pipeline performance}. 

In the GSM8K dataset, \textbf{Self-Refine V1} is the most accurate approach (86.2\%) but uses 7.4$\times$ fewer tokens compared to Pipeline V2. For math problems, two roles outperform five roles by two calls. But for HumanEval, the same \textbf{V1} drops its performance down to 66.5\%, which is 29.8 percentage points behind Direct by task-aware refactoring,

% -----------------------------------------------------------------------
% TABLE 1 DATA VERIFICATION
% Direct:        GSM8K 75.0% 295tok 4.7s  | HE 96.3% 466tok 7.1s
% Pipeline V1:   GSM8K 45.0% 8688tok 57.5s (200 problems dagger)
%                HE    56.0% 9726tok 920s  (50 problems ddagger)
% Pipeline V2:   GSM8K 82.0% 3098tok 30.4s | HE 81.7% 3881tok 35.4s
% SR V1:         GSM8K 86.2% 416tok  6.5s  (500 problems, GOOD)
%                HE    66.5% 1260tok 15.5s (164 problems, BAD)
% SR V2:         GSM8K N/A   (not run, V1 was sufficient)
%                HE    95.1% 453tok  6.3s  (164 problems)
% -----------------------------------------------------------------------

\begin{table}[t]
\centering
\caption{Main results across all methods and benchmarks.
  $\dagger$: 200-problem sample.
  $\ddagger$: 50-problem
}
\label{tab:main}
\setlength{\tabcolsep}{5pt}
\begin{tabular}{lcccc}
\toprule
\textbf{Method} & \textbf{GSM8K} & \textbf{HumanEval}
  & \textbf{Avg.\ Tokens} & \textbf{Avg.\ Latency} \\
\midrule
Direct
  & 75.0\%
  & \textbf{96.3\%}
  & 295\,/\,466
  & 4.7s\,/\,7.1s \\
Pipeline V1 (JSON)
  & 45.0\%$^{\dagger}$
  & 56.0\%$^{\ddagger}$
  & 8{,}688\,/\,9{,}726
  & 57.5s\,/\,920s \\
Pipeline V2 (Plain Text)
  & 82.0\%
  & 81.7\%
  & 3{,}098\,/\,3{,}881
  & 30.4s\,/\,35.4s \\
Self-Refine V1 (ungated)
  & \textbf{86.2\%}
  & 66.5\%
  & 416\,/\,1{,}260
  & 6.5s\,/\,15.5s \\
Self-Refine V2 (gated, HE only)
  & N/A
  & 95.1\%
  & ---\,/\,453
  & ---\,/\,6.3s \\
\bottomrule
\end{tabular}
\end{table}

\subsection{Token Efficiency}

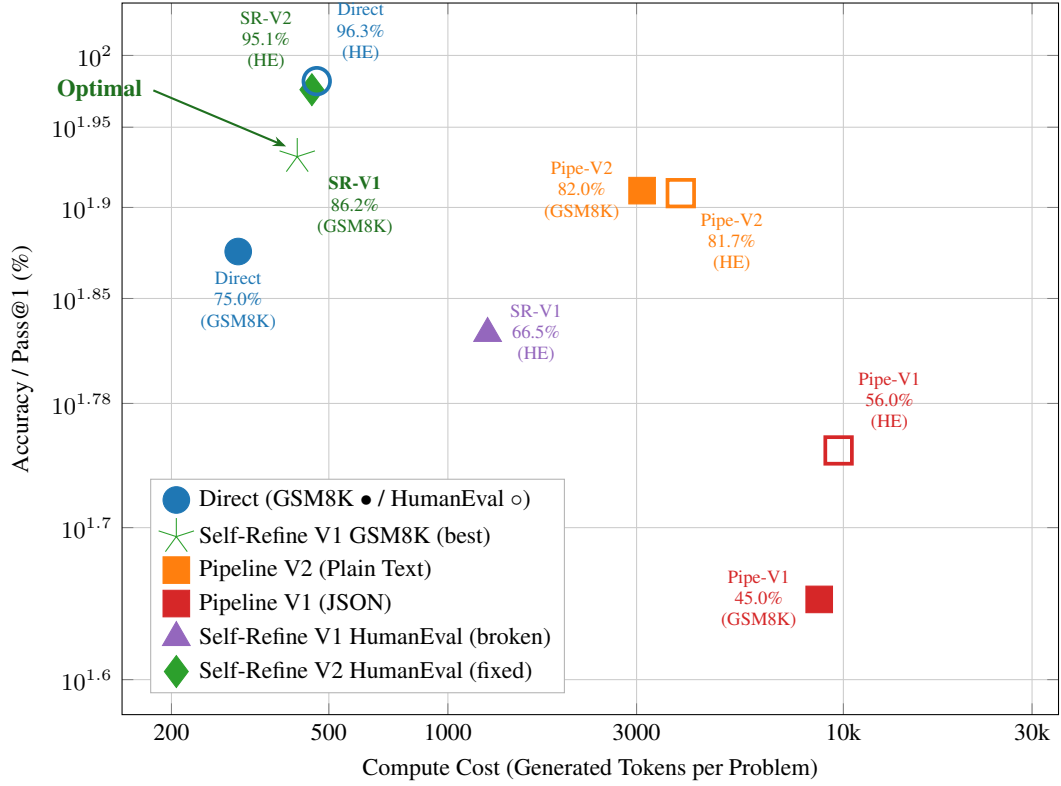
\begin{figure}[t]
\centering
\begin{tikzpicture}
\begin{loglogaxis}[
  width=1.0\linewidth,
  height=11cm,
  xlabel={Compute Cost (Generated Tokens per Problem)},
  ylabel={Accuracy / Pass@1 (\%)},
  xmin=150, xmax=35000,
  ymin=38, ymax=108,
  xtick={200,500,1000,3000,10000,30000},
  xticklabels={200,500,1000,3000,10k,30k},
  ytick={40,50,60,70,80,90,100},
  grid=both,
  grid style={line width=0.3pt, draw=gray!25},
  major grid style={line width=0.4pt, draw=gray!40},
  legend style={
    at={(0.03,0.03)},
    anchor=south west,
    font=\small,
    fill=white,
    draw=gray!60,
    row sep=1pt,
  },
  legend cell align={left},
  tick label style={font=\small},
  label style={font=\small},
  clip=false,
]

% ---- GSM8K points (solid fill) ----

% Direct GSM8K
\addplot[only marks, mark=*, mark size=5pt, color=myblue]
  coordinates {(295, 75.0)};
\addlegendentry{Direct (GSM8K $\bullet$ / HumanEval $\circ$)}

% Self-Refine V1 GSM8K — GOOD result (best accuracy)
\addplot[only marks, mark=star, mark size=7pt, color=mygreen,
         mark options={fill=mygreen}]
  coordinates {(416, 86.2)};
\addlegendentry{Self-Refine V1 GSM8K (best)}

% Pipeline V2 GSM8K
\addplot[only marks, mark=square*, mark size=5pt, color=myorange]
  coordinates {(3098, 82.0)};
\addlegendentry{Pipeline V2 (Plain Text)}

% Pipeline V1 GSM8K
\addplot[only marks, mark=square*, mark size=5pt, color=myred]
  coordinates {(8688, 45.0)};
\addlegendentry{Pipeline V1 (JSON)}

% Self-Refine V1 HumanEval — BAD result
\addplot[only marks, mark=triangle*, mark size=6pt, color=mypurple,
         mark options={fill=mypurple}]
  coordinates {(1260, 66.5)};
\addlegendentry{Self-Refine V1 HumanEval (broken)}

% Self-Refine V2 HumanEval — fixed result (diamond)
\addplot[only marks, mark=diamond*, mark size=6pt, color=mygreen,
         mark options={fill=mygreen}]
  coordinates {(453, 95.1)};
\addlegendentry{Self-Refine V2 HumanEval (fixed)}

% ---- HumanEval points (hollow / open) ----

% Direct HumanEval
\addplot[only marks, mark=o, mark size=5pt, color=myblue,
         mark options={draw=myblue, fill=white, line width=1.4pt},
         forget plot]
  coordinates {(466, 96.3)};

% Pipeline V2 HumanEval
\addplot[only marks, mark=square, mark size=5pt, color=myorange,
         mark options={draw=myorange, fill=white, line width=1.4pt},
         forget plot]
  coordinates {(3881, 81.7)};

% Pipeline V1 HumanEval
\addplot[only marks, mark=square, mark size=5pt, color=myred,
         mark options={draw=myred, fill=white, line width=1.4pt},
         forget plot]
  coordinates {(9726, 56.0)};

% ---- Labels for GSM8K points ----
\node[font=\scriptsize, align=center, anchor=north, text=myblue]
  at (axis cs:295, 75.0) [yshift=-4pt] {Direct\\75.0\%\\(GSM8K)};

\node[font=\scriptsize, align=center, anchor=north west, text=mygreen!70!black]
  at (axis cs:416, 86.2) [xshift=4pt, yshift=-4pt] {\textbf{SR-V1}\\86.2\%\\(GSM8K)};

\node[font=\scriptsize, align=center, anchor=east, text=myorange]
  at (axis cs:3098, 82.0) [xshift=-5pt] {Pipe-V2\\82.0\%\\(GSM8K)};

\node[font=\scriptsize, align=center, anchor=east, text=myred]
  at (axis cs:8688, 45.0) [xshift=-5pt] {Pipe-V1\\45.0\%\\(GSM8K)};

% ---- Labels for HumanEval points ----
\node[font=\scriptsize, align=center, anchor=south west, text=myblue]
  at (axis cs:466, 96.3) [xshift=4pt, yshift=4pt] {Direct\\96.3\%\\(HE)};

\node[font=\scriptsize, align=center, anchor=south east, text=mygreen!70!black]
  at (axis cs:453, 95.1) [xshift=-4pt, yshift=4pt] {SR-V2\\95.1\%\\(HE)};

\node[font=\scriptsize, align=center, anchor=west, text=mypurple]
  at (axis cs:1260, 66.5) [xshift=5pt] {SR-V1\\66.5\%\\(HE)};

\node[font=\scriptsize, align=center, anchor=north west, text=myorange]
  at (axis cs:3881, 81.7) [xshift=4pt, yshift=-4pt] {Pipe-V2\\81.7\%\\(HE)};

\node[font=\scriptsize, align=center, anchor=south west, text=myred]
  at (axis cs:9726, 56.0) [xshift=4pt, yshift=4pt] {Pipe-V1\\56.0\%\\(HE)};

% ---- "Optimal" annotation arrow ----
\draw[-{Stealth[length=5pt]}, thick, mygreen!70!black]
  (axis cs:180, 95) -- (axis cs:390, 87.5);
\node[font=\small\bfseries, color=mygreen!70!black, anchor=east]
  at (axis cs:178, 95) {Optimal};

\end{loglogaxis}
\end{tikzpicture}
\caption{Accuracy vs.\ average tokens per problem (log scale).
  }
\label{fig:pareto}
\end{figure}

Self-Refine V1 consumes 1.4$\times$ the number of tokens as Direct on GSM8K but performs the best in terms of accuracy on GSM8K. For HumanEval, Self-Refine V2 consumes only slightly fewer tokens compared to Direct (453 vs.\ 466 tokens) because the confidence gate does not execute the second call (Call~2) on 163 of 164 examples. Self-Refine V1 consumes 2.7$\times$ the number of tokens consumed by Direct but scores 29.8 points lower than Direct--the inverse of what would be expected. The Pipeline V1 consumes close to 30$\times$ more tokens than Direct but attains the lowest accuracy for both tasks.Figure~\ref{fig:pareto} visualizes the accuracy vs.\ token-cost tradeoff across all methods.

%------ Figure 1: Pareto scatter (Accuracy vs Tokens) --------------------

\subsection{Self-Refine Detailed Analysis}
\label{sec:sr_detail}

Table~\ref{tab:sr_detail} breaks down the call-by-call contribution of
self-refinement across all evaluated conditions.

% -----------------------------------------------------------------------
% TABLE 2 DATA VERIFICATION
% SR-V1 GSM8K:      Call1 85.4% -> Final 86.2%  Fixed:5 Hurt:1
%                   TokC1:376  TokC2:39
% SR-V1 HumanEval:  Call1 95.1% -> Final 66.5%  Fixed:4 Hurt:51  Skip:0
%                   TokC1:452  TokC2:807
% SR-V2 HumanEval:  Call1 95.1% -> Final 95.1%  Fixed:0 Hurt:0   Skip:163
%                   TokC1:453  TokC2:~0
% -----------------------------------------------------------------------

\begin{table}[t]
\centering
\caption{Self-refinement call-by-call analysis.
  (Call~2, SR-V2 was evaluated on HumanEval only.)}
\label{tab:sr_detail}
\setlength{\tabcolsep}{4pt}
\begin{tabular}{llccccccc}
\toprule
\textbf{Variant} & \textbf{Benchmark}
  & \textbf{Call~1}
  & \textbf{Final}
  & \textbf{Fixed}
  & \textbf{Hurt}
  & \textbf{Skipped}
  & \textbf{Tok C1}
  & \textbf{Tok C2} \\
\midrule
SR-V1 (ungated) & GSM8K 
  & 85.4\% & 86.2\% & 5 & 1 & --- & 376 & 39 \\
SR-V1 (ungated) & HumanEval 
  & 95.1\% & 66.5\% & 4 & 51 & 0 & 452 & 807 \\
\midrule
SR-V2 (gated)   & HumanEval 
  & 95.1\% & 95.1\% & 0 & 0 & 163 & 453 & $\approx$0 \\
\bottomrule
\end{tabular}
\end{table}

For GSM8K, Self-Refine V1 Call~2 is useful: 5 problems get corrected for a price of 1 regression; overall improvement of 4 problems ($+$0.8\%). The average number of 39 tokens generated by Call~2 indicates the model is affirming correct solutions instead of writing them anew---the most effective use of the model’s capacity.

For HumanEval, Self-Refine V1 Call~2 is catastrophic: 95.1\% drops to 66.5\% because 4 problems get fixed and 51 break---a 28.6\% degradation. Self-Refine V2 resolves the issue, as Call~2 skips on 163 out of 164 problems through the confidence gate, and the only remaining problem results in \texttt{CORRECT} without modifying the original code, preserving the initial accuracy of 95.1\%.

\subsection{Pipeline V1 Runtime}
\label{sec:runtime}

Practical remark: In order to complete the entire HumanEval set, we needed to run for at least 13 hours to solve only the first 50 problems of V1 on the pipeline and an average time delay of 920 seconds per problem. It will take about 42 hours to solve all 164 problems. This runtime is completely impractical for any real deployment.

%==========================================================================
\section{Analysis}
%==========================================================================

\subsection{Why JSON Communication Fails}

This Pipeline V1 failure follows a well-documented pattern. Each role is asked to produce a structured JSON object. A 7B model producing a response according to a complicated system prompt and valid JSON at the same time will produce malformed output---missing brackets, broken escaping, truncated content, and additional commentary not part of the JSON object.

\begin{figure}[t]
\centering
\begin{tikzpicture}[
  node distance = 0.55cm and 1.1cm,
  box/.style = {
    rectangle, rounded corners=3pt,
    minimum width=2.3cm, minimum height=0.7cm,
    text centered, font=\small,
    draw, thick,
  },
  role/.style  = {box, fill=myblue!15,  draw=myblue!60},
  error/.style = {box, fill=myred!20, draw=myred!70, minimum width=2.6cm},
  ok/.style    = {box, fill=mygreen!20, draw=mygreen!70, minimum width=2.6cm},
  arr/.style   = {-{Stealth[length=4pt]}, thick},
  errarr/.style= {-{Stealth[length=4pt]}, thick, myred},
  okarr/.style = {-{Stealth[length=4pt]}, thick, mygreen!70!black},
  label/.style = {font=\small\bfseries},
  sublabel/.style = {font=\scriptsize\itshape},
]

%% --- LEFT SIDE: Pipeline V1 (JSON) ---
\node[label] (v1title)
  {\textcolor{myred}{Pipeline V1 --- JSON Communication}};

\node[role, below=0.45cm of v1title] (d1) {Darbari};
\node[error, below=0.42cm of d1] (e1) {\texttt{\{malformed\,JSON\}}};
\node[role, below=0.42cm of e1] (m1) {Majumdar};
\node[error, below=0.42cm of m1] (e2) {garbled context};
\node[role, below=0.42cm of e2] (s1) {Sainik};
\node[error, below=0.42cm of s1] (wrong) {\textbf{Wrong answer}};

\draw[arr, myred!70] (d1) --
  node[right, font=\scriptsize, text=myred!80]
    {parse fail $\to$ raw text} (e1);
\draw[errarr] (e1) -- (m1);
\draw[arr, myred!70] (m1) --
  node[right, font=\scriptsize, text=myred!80]
    {context lost} (e2);
\draw[errarr] (e2) -- (s1);
\draw[errarr] (s1) -- (wrong);

\node[sublabel, below=0.15cm of wrong, text=myred!80]
  {45.0\% GSM8K\,/\,56.0\% HumanEval};

%% --- RIGHT SIDE: Pipeline V2 (Plain Text) ---
\node[label, right=4.4cm of v1title] (v2title)
  {\textcolor{mygreen!70!black}{Pipeline V2 --- Plain Text}};

\node[role, below=0.45cm of v2title] (d2) {Darbari};
\node[ok,   below=0.42cm of d2] (t1) {plain text analysis};
\node[role, below=0.42cm of t1] (m2) {Majumdar};
\node[ok,   below=0.42cm of m2] (t2) {numbered plan};
\node[role, below=0.42cm of t2] (s2) {Sainik};
\node[ok,   below=0.42cm of s2] (correct) {\textbf{Correct answer}};

\draw[okarr] (d2) --
  node[right, font=\scriptsize, text=mygreen!70!black]
    {passed cleanly} (t1);
\draw[okarr] (t1) -- (m2);
\draw[okarr] (m2) --
  node[right, font=\scriptsize, text=mygreen!70!black]
    {full context maintained} (t2);
\draw[okarr] (t2) -- (s2);
\draw[okarr] (s2) -- (correct);

\node[sublabel, below=0.15cm of correct, text=mygreen!70!black]
  {82.0\% GSM8K\,/\,81.7\% HumanEval};

%% --- Cascade annotation ---
\draw[decorate, decoration={brace, amplitude=5pt, mirror},
      thick, myred!60]
  ([xshift=-3pt]d1.north west) -- ([xshift=-3pt]wrong.south west)
  node[midway, left=8pt, font=\scriptsize, align=right, text=myred!80]
  {Cascading\\failure};

\end{tikzpicture}
\caption{JSON communication V/S Plain Text Communication}
\label{fig:pipeline_compare}
\end{figure}
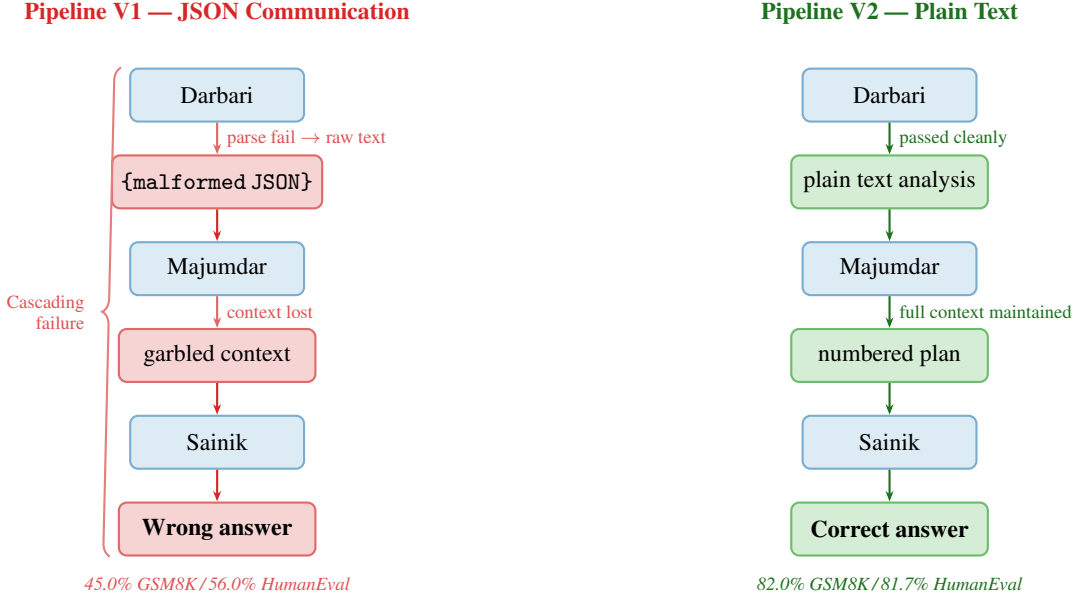

If malformed JSON is encountered by the engine, it defaults to passing plain text to the next role downstream. The next role then gets a prompt which includes JSON fragments intermixed with parts of reasoning in natural language and without context for the initial question anymore. By the third or fourth role, the chain has strayed far from its starting point. Figure~\ref{fig:pipeline_compare} highlights this process.

%------ Figure 3: Pipeline Communication Format comparison (TikZ diagram) -

The fix achieved by Pipeline V2 supports this explanation. Roles generating numbered steps and natural language instead of JSON lead to successful responses. The same 7B model generating malformed JSON 30--40\% of the time produces valid natural language responses almost all the time.

\textbf{Implications for practitioners:} Avoid using structured JSON output requirements for sub-10B models. Stick to plain text with minimal formatting.

\subsection{Why Self-Refine V1 Fails on Code (And How V2 Fixes It)}

One of the most noteworthy outcomes of this evaluation is Self-Refine V1 on HumanEval: starting from 95.1\% initial accuracy after Call~1, the generic ungated Call~2 step reduced final accuracy to 66.5\%, after fixing 4 errors and introducing 51. Note that the same V1 implementation worked well on GSM8K ($+$0.8\% net). The failure is specific to code tasks. Three contributing factors have been identified.

%------ Figure 2: Self-Refine Call-by-Call on HumanEval ------------------
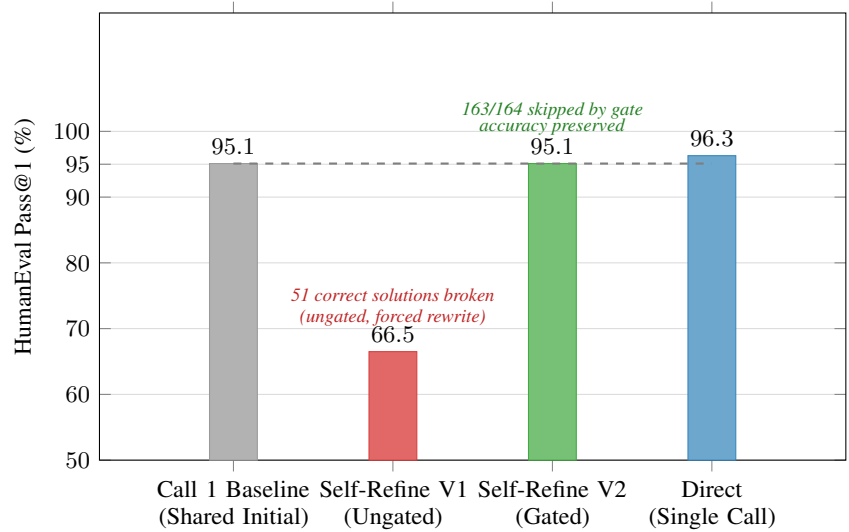
\begin{figure}[t]
\centering
\begin{tikzpicture}
\begin{axis}[
  width=0.82\linewidth,
  height=7.5cm,
  ybar,
  bar width=18pt,
  xtick={1,2,3,4},
  xticklabels={
    {Call 1 Baseline\\(Shared Initial)},
    {Self-Refine V1\\(Ungated)},
    {Self-Refine V2\\(Gated)},
    {Direct\\(Single Call)}
  },
  xticklabel style={font=\small, text width=2.6cm, align=center},
  ymin=50, ymax=118,
  ytick={50,60,70,80,90,95,100},
  ylabel={HumanEval Pass@1 (\%)},
  ylabel style={font=\small},
  nodes near coords,
  nodes near coords align={vertical},
  nodes near coords style={font=\small\bfseries,
    /pgf/number format/.cd, fixed, precision=1},
  enlarge x limits=0.28,
  ymajorgrids=true,
  grid style={line width=0.3pt, draw=gray!30},
  tick label style={font=\small},
  clip=false,
  axis on top=false,
]
% Baseline
\addplot[fill=mygray!60, draw=mygray!80, bar shift=0pt] coordinates {
  (1, 95.1)
};
% V1 -- bad
\addplot[fill=myred!70, draw=myred!90, bar shift=0pt] coordinates {
  (2, 66.5)
};
% V2 -- good
\addplot[fill=mygreen!65, draw=mygreen!90, bar shift=0pt] coordinates {
  (3, 95.1)
};
% Direct
\addplot[fill=myblue!65, draw=myblue!90, bar shift=0pt] coordinates {
  (4, 96.3)
};

% Reference line at 95.1
\draw[dashed, thick, mygray]
  (axis cs:1, 95.1) -- (axis cs:4, 95.1);

% Annotations
\node[font=\scriptsize, text=myred!90!black, align=center, anchor=south]
  at (axis cs:2, 69)
  {\textit{51 correct solutions broken}\\
   \textit{(ungated, forced rewrite)}};

\node[font=\scriptsize, text=mygreen!80!black, align=center, anchor=south]
  at (axis cs:3, 98)
  {\textit{163/164 skipped by gate}\\[-2pt]
   \textit{accuracy preserved}};

\end{axis}
\end{tikzpicture}
\caption{Effect of the self-refinement Call~2 step on HumanEval pass@1.
 }
\label{fig:selfrefine_he}
\end{figure}

\textbf{Cause 1: Code confidence gate failed.} The V1 gate checked for uncertainty markers (``I think'', ``might be'', ``not certain'') within the entire response. Python code frequently contains these phrases in comments and docstrings, causing the gate to classify all 164 responses as uncertain and running Call~2 unconditionally.

\textbf{Reason 2: Urgency causes superfluous rewriting.} Call~2 was asked to ``always present the final function enclosed by code blocks.'' The model would even rewrite the perfectly working function out of habit, resulting in bugs being added to working code.

\textbf{Reason 3: Too high initial accuracy approaching ceiling.} With 95.1\% initial accuracy, 156 out of 164 examples are working correctly. Applying Call~2 to all 164 examples will result in 164 chances of breaking working code but only 8 chances of fixing incorrect code.

All of these problems are fixed in Self-Refine V2: a gate specific for code that detects the end of code block instead of detecting uncertainty marker; a cautious prompt asking the model to respond with \texttt{CORRECT} whenever there are no bugs detected; and returning the initial response from Call~1 output when Call~2 confirms correctness.

\subsection{Ablation: What Actually Matters in Self-Refine}

Table~\ref{tab:ablation} presents the HumanEval ablation isolating each design decision. Variant~A (no gate, non-coercive prompt) achieves 96.3\%---\emph{marginally above} the Call~1 baseline of 95.1\%. This is unexpected: V1 (also no gate) scored 66.5\%. The difference is the Call~2 prompt. Removing the forced-rewrite instruction eliminates the 51-problem regression entirely.

Variant~B (gate on, conservative prompt) scores 95.1\%, identical to Call~1. The gate skips 163 problems; the one that reaches Call~2 returns \texttt{CORRECT} and the original code.

Variant~C (gate on, force-rewrite) scores 94.5\%, breaking 1 problem---confirming that the force-rewrite instruction is harmful even when it fires rarely.

\textbf{Counterintuitive finding:} The no-gate approach (Variant~A) works better than the gate versions, since the gate prevents two fixes that Variant~A achieves. The best choice for near-ceiling tasks is no gate with non-coercive prompts. For tasks with improvement potential (like GSM8K), the gate is not critical, as Call~2 helps and rarely hurts on math problems.

\begin{table}[t]
\centering
\caption{Self-Refine ablation on HumanEval.}
\label{tab:ablation}
\setlength{\tabcolsep}{6pt}
\begin{tabular}{lcccc}
\toprule
\textbf{Variant} & \textbf{Pass@1} & \textbf{Fixed}
  & \textbf{Hurt} & \textbf{Skipped} \\
\midrule
Call~1 baseline (shared)
  & 95.1\% & --- & --- & --- \\
\midrule
Self-Refine V1 (no gate $+$ force-rewrite)
  & 66.5\% & 4 & 51 & 0 \\
Variant A: no gate $+$ generic prompt
  & \textbf{96.3\%} & 2 & 0 & 0 \\
Variant B: gate $+$ conservative prompt
  & 95.1\% & 0 & 0 & 163 \\
Variant C: gate $+$ force-rewrite
  & 94.5\% & 0 & 1 & 163 \\
\midrule
Direct (reference)
  & 96.3\% & --- & --- & --- \\
\bottomrule
\end{tabular}
\end{table}

\subsection{Task Type Determines Optimal Strategy}

By summing up all observations, we can draw the following conclusion. In cases where the direct accuracy still needs some improvement (GSM8K at 75\%), the process of self-refinement always helps the model improve. In situations where the model already demonstrates maximum direct accuracy possible (HumanEval at 96.3\%), self-refinement does not help but costs extra effort.

\textbf{Recommendation for practice:} determine the direct accuracy first. If it is under 85\%, self-refinement should help. If direct accuracy is above 90\%, self-refinement will cost extra effort but yield nothing.

%==========================================================================
\section{Discussion}
%==========================================================================

\subsection{Why Pipelines Were Expected to Help}

The logic for the multirole pipeline approach appears reasonable. Dividing the complex task into planning, acting, and verifying will result in an optimized cognitive workload distribution, specialization, and improved error checking. This is possible only if each of the roles is able to accomplish its task.

The unreliable structure assumption failure occurs for 7B models. GPT-4-class models reliably generate a correct JSON structure when prompted to create one in a complex system prompt. But the 7B model cannot do so. Once the assumption of reliable output structuring breaks down, it ruins everything, since the context in each successive stage is broken by default.

\subsection{The Plain Text Recovery}

This 37 percentage point improvement relative to plain text communication also has important engineering implications. Engineers who implement multiagent systems for local models need to focus their communication on the strength of small models, which is natural language descriptions, rather than that of large models, which is structured formats.

However, it must be pointed out that plain text role communication means one capability of structured format is lost: parsing and validating the role outputs programmatically. Our results suggest this is a worthwhile tradeoff for local models.

\subsection{The Self-Refine Tradeoff}

Self-Refine V1 provides the most accurate results on GSM8K with the least amount of tokens for all algorithms that outperform Direct. The main utility of self-refinement lies in increasing the baseline accuracy---the 5 mistakes in Call~2 would have been incorrect solutions otherwise---instead of transforming the distribution of accuracy results drastically. The problem with V1's inability to solve HumanEval questions does not reside within the self-refinement framework itself; rather, the ablation analysis proves that using a noncoercive prompt (Variant~A) resolves the entire issue. The failure was in the specific forced-rewrite instruction.

\subsection{Limitations}

We have certain limitations in our evaluation. For example, we evaluate only one family of models, namely Qwen2.5-7B-Instruct. Our results may vary for other families such as Llama, Mistral, or Phi families. We evaluated two benchmarking sets, but our results for multi-hop reasoning, factual QA, or long-form generation might differ from those in our study. Pipeline V1 HumanEval evaluation is done using 50 tasks because of the high latency of 920 seconds. Our study does not include evaluation for larger models, 13B or 34B.
%==========================================================================
\section{Conclusion}
%==========================================================================

A multi-agent pipeline comprising five roles was tested against direct prompting and self-refinement methods on the local version of Qwen2.5-7B. JSON-based communication using the multi-agent pipeline reduces accuracy by more than 30 percent on both datasets compared to plain text communication which restores nearly all these lost accuracies. The two-step refinement method using V1 is better than the static pipeline on math by requiring 7.4$\times$ less token usage. However, the use of the V1 on HumanEval decreases accuracy by 29.8 percent since the rewrite prompt causes the model to write broken code. Task-specific refinement using V2 brings back the accuracy to 95.1\%.

The main lesson from this experiment is that \textbf{complex design cannot compensate for poor implementation}. The five-part pipeline with incorrect communication protocol underperforms compared to one call from a single model. Similarly, two-call self-refinement with the wrong prompting setup underperforms compared to its own first call. Ensuring a proper implementation in terms of communication protocol, prompt design, confidence gating, and answer selection trumps architectural complexity.

Practitioners using LLMs locally on 7B models: Start with direct prompts as your baseline; incorporate self-refinement for tasks where direct prompt accuracy is less than 85\%; and do not use structured JSON-based communication in multi-step pipelines.

\begin{ack}
The authors thank the open-source communities behind \texttt{llama.cpp} and the Qwen team for releasing model weights.
\end{ack}

\bibliographystyle{plain}

\begin{thebibliography}{99}

\bibitem{bai2025qwen}
J.~Bai et al.
\newblock Qwen2.5 Technical Report.
\newblock {\em arXiv preprint arXiv:2412.15115}, 2025.

\bibitem{chen2021humaneval}
M.~Chen et al.
\newblock Evaluating Large Language Models Trained on Code.
\newblock {\em arXiv preprint arXiv:2107.03374}, 2021.

\bibitem{cobbe2021gsm8k}
K.~Cobbe et al.
\newblock Training Verifiers to Solve Math Word Problems.
\newblock {\em arXiv preprint arXiv:2110.14168}, 2021.

\bibitem{hong2024metagpt}
S.~Hong et al.
\newblock MetaGPT: Meta Programming for Multi-Agent Collaborative Framework.
\newblock In {\em ICLR}, 2024.

\bibitem{huang2023agentcoder}
D.~Huang et al.
\newblock AgentCoder: Multi-Agent-based Code Generation with Iterative Testing and Optimisation.
\newblock {\em arXiv preprint arXiv:2312.13010}, 2023.

\bibitem{kojima2022large}
T.~Kojima et al.
\newblock Large Language Models are Zero-Shot Reasoners.
\newblock In {\em NeurIPS}, 2022.

\bibitem{madaan2023selfrefine}
A.~Madaan et al.
\newblock Self-Refine: Iterative Refinement with Self-Feedback.
\newblock In {\em NeurIPS}, 2023.

\bibitem{qian2024chatdev}
C.~Qian et al.
\newblock ChatDev: Communicative Agents for Software Development.
\newblock In {\em ACL}, 2024.

\bibitem{shinn2023reflexion}
N.~Shinn et al.
\newblock Reflexion: Language Agents with Verbal Reinforcement Learning.
\newblock In {\em NeurIPS}, 2023.

\bibitem{crewai2024}
J.~Moura.
\newblock CrewAI: Framework for Orchestrating Role-Playing Autonomous AI Agents.
\newblock \url{https://github.com/joaomdmoura/crewAI}, 2024.

\bibitem{wei2022chain}
J.~Wei et al.
\newblock Chain-of-Thought Prompting Elicits Reasoning in Large Language Models.
\newblock In {\em NeurIPS}, 2022.

\bibitem{wu2023autogen}
Q.~Wu et al.
\newblock AutoGen: Enabling Next-Gen LLM Applications via Multi-Agent Conversation.
\newblock {\em arXiv preprint arXiv:2308.08155}, 2023.

\end{thebibliography}

%\newpage
%\input{checklist.tex}

\end{document}